\RequirePackage{fix-cm}
\documentclass[onecolumn]{svjour3}          % twocolumn

\smartqed  % flush right qed marks, e.g. at end of proof
\usepackage{color}
\usepackage{booktabs}
\usepackage[flushleft]{threeparttable}
\usepackage{graphicx}
\usepackage{setspace}
\usepackage[left=3cm,right=3cm,top=3cm,bottom=3cm]{geometry}

\usepackage{array,tabularx,booktabs}
\usepackage{makecell}
\usepackage{boldline}
\usepackage{rotating}
\usepackage{adjustbox}
\usepackage{lscape}
\usepackage{amssymb,amsfonts,amsbsy,epsfig}
\setcellgapes{3pt}
\usepackage{flushend}
\usepackage{epstopdf}
\usepackage{framed}
\usepackage{amsmath}
\usepackage{marvosym}
\usepackage{arabtex}
\usepackage{utf8}
\setcode{utf8}
\title{
Sentiment analysis for Arabic language: A brief survey of approaches and techniques
}
\begin{document}
\author{Mo'ath Alrefai \and Hossam Faris \and and Ibrahim Aljarah}

 \institute{M. Alrefai
 		   \and
             H. Faris \Letter
            \and 
            I. Aljarah\at
            King Abdullah II School for Information Technology, The University of Jordan, Amman, Jordan \\
            \email{\{hossam.faris,i.aljarah\}@ju.edu.jo}
 \\
            \email{maa8141201@fgs.ju.edu.jo}
            % do you have official email address @ ju.edu.jo ? if so, please add it first. -- Here is my email MAA8141201@FGS.JU.EDU.JO , Moath!
 }
% \author{ \and \and }

% \institute{
		   % \and
            % \Letter
           % \and 
           % \at
            % \\
           % \email{}
% \\
           % \email{}
           % % do you have official email address @ ju.edu.jo ? if so, please add it first. -- Here is my email MAA8141201@FGS.JU.EDU.JO , Moath!
% }
\date{Received: date / Accepted: date}
\maketitle

%%%%%%%%%%%%%%%%%%%%%%%%%%%%%%%%%%%%%%%%%%%%%%%%%%%%%%%%%%%%%%%%%%%%%%%%%%%%%%%%
%%%%%%%%%%%%%%%%%%%%%%%%%%%%%%%%%%%%%%%%%%%%%%%%%%%%%%%%%%%%%%%%%%%%%%%%%%%%%%%%
\begin{abstract}

With the emergence of Web 2.0 technology and the expansion of on-line social networks, current Internet users have the ability to add their reviews, ratings and opinions on social media and on commercial and news web sites. Sentiment analysis aims to classify these reviews reviews in an automatic way. In the literature, there are numerous approaches proposed for automatic sentiment analysis for different language contexts. Each language has its own properties that makes the sentiment analysis more challenging. In this regard, this work presents a comprehensive survey of existing Arabic sentiment analysis studies, and covers the various approaches and techniques proposed in the literature. Moreover, we highlight the main difficulties and challenges of Arabic sentiment analysis, and the proposed techniques in literature to overcome these barriers.

\end{abstract}

%%%%%%%%%%%%%%%%%%%%%%%%%%%%%%%%%%%%%%%%%%%%%%%%%%%%%%%%%%%%%%%%%%%%%%%%%%%%%%%%
\section{Introduction}
With the emergence of Web 2.0 technology and rising number of social media websites and web forums, current Internet users have the ability to add their reviews, ratings or opinions on web pages whether it is a commercial website or news website or any other website. Furthermore, people by nature often consult discussion forums before making investment decision by asking friends about their opinions, and they read consumers’ reviews before making a big purchase like an electrical device or a car. So, the consumers tend to trust other consumers’ opinions rather than advertisements, and this lead to a better decision for the consumer especially in case when he doesn’t have a lot of experience or knowledge about the target product \cite{c1,shukri2015twitter,qaisi2016twitter,hakh2017online}. Not only from consumers side, the marketers will try to get the bad reviews about their products and report the issues to their developers and designers \cite{c5} so they can meet the public demands.

Extracting manually good or bad reviews for a specific product is extremely cumbersome and time consuming, and this in turn led to the emergence of a new technique for extracting the opinions related to a specific topic which is called Sentiment Analysis (SA). SA refers to use the Natural Language Processing (NLP) and text analysis to extract the sentiment from a given text that is related to specific topic \cite{c2,ghadeer2017enhancing}. Sharda et. al. in \cite{c1} defined SA as ``a technique used to detect favorable and unfavorable opinions toward specific products and services using large numbers of textual data sources''. By using sentiment analysis we can answer some important questions like this "What do the consumers say about this device or this service" by collecting the reviews about the device or the service and then digging into opinions using automatic sentiment analysis tool to get the answer. 

Arabic is currently ranked as the forth language used in the web, and there are about 168 million of Arabic Internet user \cite{c3}. Most of sentiment analysis works focus on English language and there is relatively less work on Arabic language compared with English. The complexity of Arabic language and the lack of the available resources of Arabic sentiment analysis like lexicons and datasets are the main obstacles in Arabic sentiment analysis \cite{c4}. In the literature on SA, there are two major approaches to build SA system, machine learning approach and lexicon-based approach, also, in some literatures we found the two approaches were combined to build a hybrid one. In this work we review the existing Arabic Sentiment Analysis studies and and cover the various approaches proposed in literature. In addition, we highlight the main challenges of Arabic Sentiment Analysis.

The rest of this work is organized as follows: section 2 provides a brief background on Sentiment Analysis approaches; section 3 talks about the main challenges in Arabic Sentiment Analysis; section 4 reviews Sentiment Analysis process; section 5 contains the conclusion and future work.

\section{Challenges of Arabic language}

Arabic is one of the six official languages of the United Nations and it is a morphologically rich language. This language has two main forms, standard and dialectal. Modern Standard Arabic (MSA) used in formal speeches and writing like book and newspapers while dialect Arabic used in informal writing specially in social media and it vary from one country to another \cite{c13}.   
Analyzing Arabic text is very complicated, the following points explain the complexity of applying sentiment analysis on Arabic language.  

\begin{itemize}

\item Every Arabic country has its own dialect that vary from other country, and people tend to use their dialect instead of using MSA.

\item The same word can be written by different users in different ways like the words that ends with Ta' marbootah (\RL{ة}), for example, (\RL{المؤثرة}) can be written as (\RL{المؤثره}).

\item The negation words can invert the meaning of the sentence, so if the sentence had a positive sentiment then after adding the negation word in the sentence then it will have had a negative sentiment. In Natural Language Processing (NLP), negation words are considered stop words and it will be removed from the sentence as a preprocessing step.

\item The same verb can be written in different ways based on the subject of the verb, singular or plural, feminine or masculine. For example, "\RL{هو يحب السيارات}" (He likes cars) and "\RL{هي تحب السيارات}" (She likes cars).

\item Some Arabic names are derived from adjectives, and the name itself does not have s sentiment while the adjective may have a sentiment. For example, the name Jameelah and the adjective Beautiful have the same form in Arabic (\RL{جميلة}).

\item Arabic speakers could use idioms to express their opinions, and the idioms have implicit opinion. For example, the idiom \RL{حسبي الله ونعم الوكيل} has a negative opinion while it does not have any negative word.
\end{itemize}

%In this research we are concerned about MS and the Jordanian dialects. The complexity here is that the NLP techniques does not deal with Arabic dialects \cite{c13} so we used other techniques to deal with these difficulties. 

\section{Sentiment analysis approaches}

Sentiment analysis is considered as text mining techniques that can be used to detect favorable and unfavorable opinions toward specific services or products \cite{c1}, and it can be applied on different areas such as brand Management, politics, healthcare, financial markets, or in e-mail filtration by prioritizing received e-mails.

According to a recent survey presented in \cite{c4}, sentiment analysis on Arabic is still in the early stages and the research was rapidly increasing in recent years, 2014, and 2015 . We have collected almost all Arabic Sentiment analysis researches from Springer, IEEE, ACM, and the references in Arabic Sentiment Analysis surveys, the collection has been done in August 2017, the following sub-sections show some statistical reports about the collected literatures and discuss the literatures based on the used approach. 

\begin{figure*}
\centering
  \includegraphics[scale=0.4]{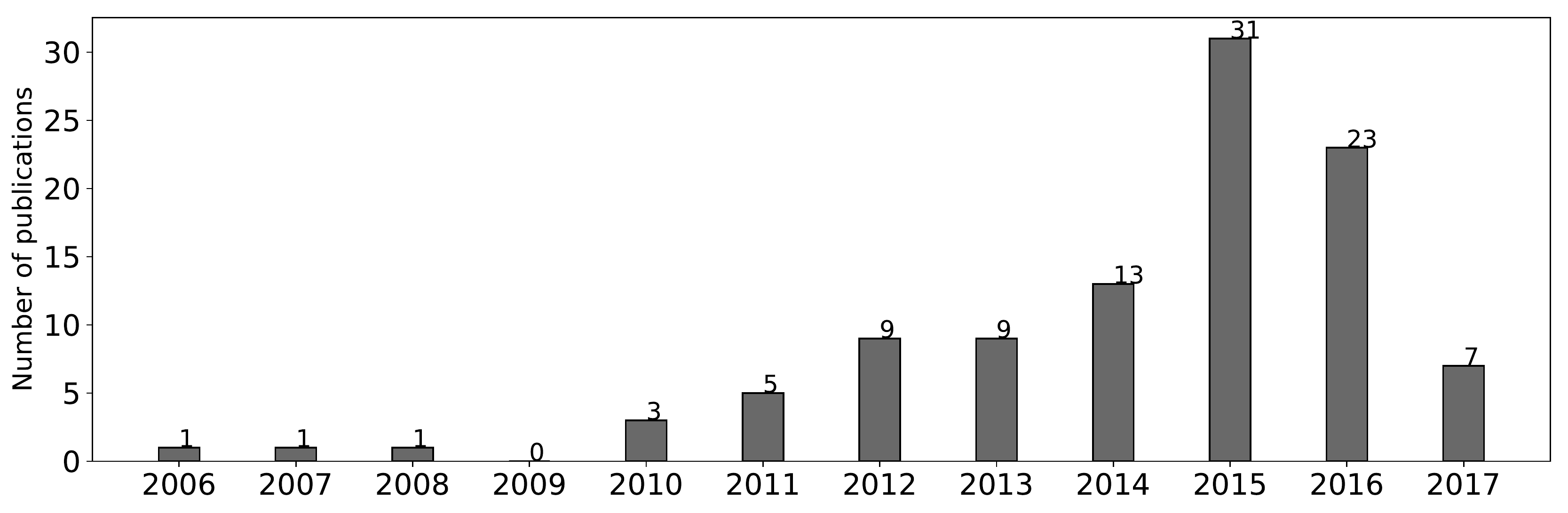}
  \caption{Count of surveyed publications of sentiment analysis for Arabic language per year}
  \label{fig:Publish_date_histogram}
\end{figure*}

%\subsection{Statistical reports}
Figure \ref{fig:Publish_date_histogram} shows the publish date histogram report for the collected researches, and as we can see, the researches started from 2006 and they have rapidly risen in recent years. And according to the survey \cite{c204}, Arabic sentiment analysis is still an open area for research. 

\begin{figure}
\centering
  \includegraphics[scale=0.7]{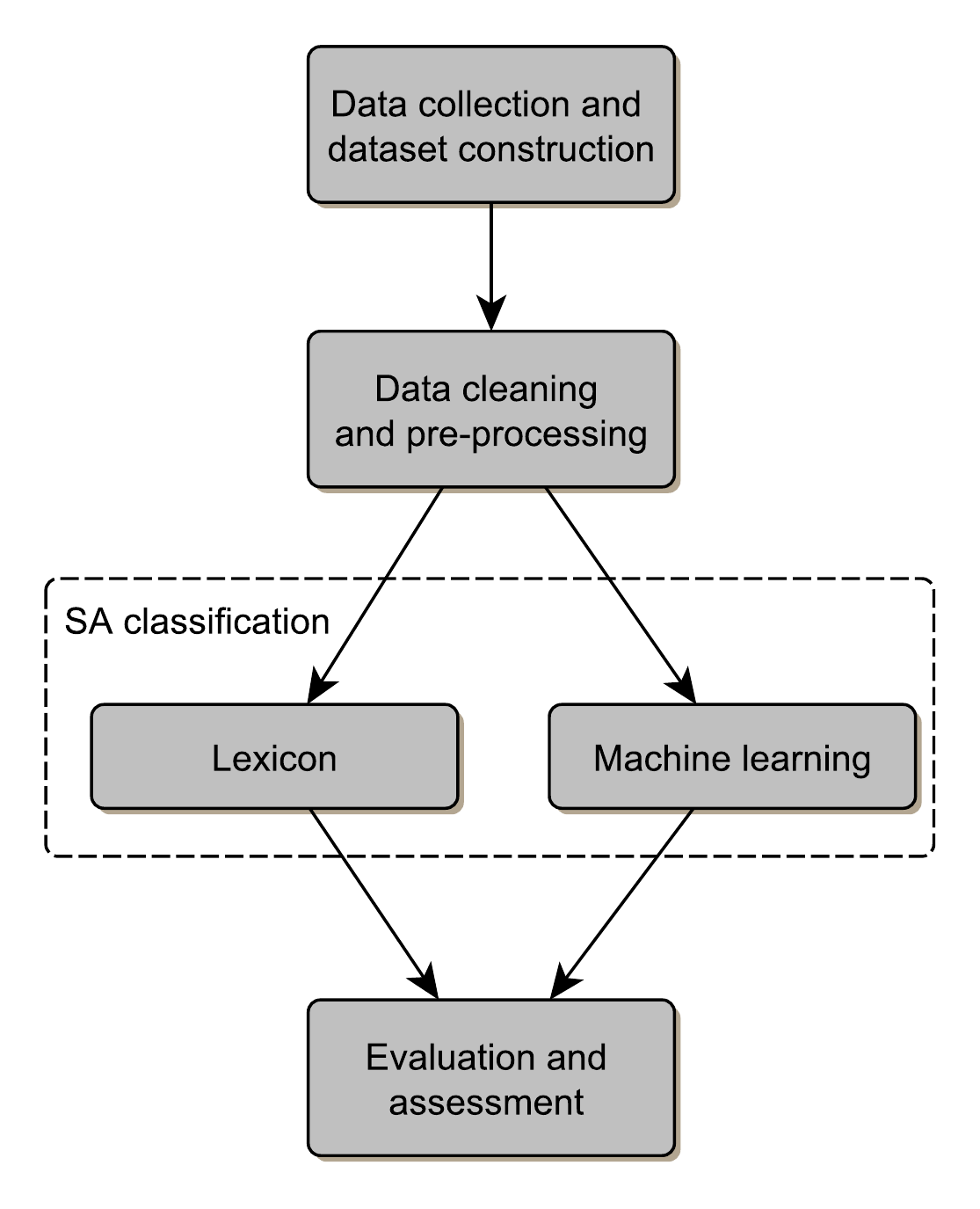}
  \caption{Flowchart of main processes in sentiment analysis approaches}
  \label{fig:SA_Process}
\end{figure}

The process of building Sentiment Analysis module in general consists of four main steps: Data Collection, pre-processing, sentiment classification, and evaluation. Figure \ref{fig:SA_Process} shows these steps.

\subsection{Datasets collection and annotation}

In the first step, the dataset is collected and annotated, and then it will be divided into training and testing datasets for Machine learning approach. For annotation process where each instance in the dataset should assign to a specific class like positive, negative, or neutral, there are two used approaches for annotation process, the first one is the manual annotation approach where two or more annotators classify each instance in the dataset manually, this approach has been used in \cite{c42,c64}. And the second approach is the crowd-sourcing approach where a crowd of Internet users classify the instances in the dataset by using a web application that is created by the authors, this approach has been used in \cite{c8,c109}.

Some works have focused on building the dataset, an example of such works was presented in \cite{c10}, where the authors have presented an Arabic corpus which was collected from different Arabic blogs and web pages. The corpus contained 250 positive and 250 negative reviews were written in MSA and it is publicly available for research purpose. Also, the work in \cite{c29} has presented an Arabic Sentiment Tweet Dataset (ASTD) which gathered from Twitter, the dataset contained about 10k tweets classified as objective and subjective (positive, negative, and mixed), and it is publicly available.

\subsection{Datasets cleaning and preparation}

In the second step of building Sentiment Analysis module, the dataset is cleaned where some instances will be removed from the dataset like removing the duplicated instances and the instances that has a long text, then after cleaning the dataset, the pre-processing operations will be applied on each instance in the dataset. Many pre-processing techniques have been used in this steps, the following points describe these techniques:
\begin{itemize}

\item Text normalization and removing repeated letters: transforming some Arabic letters to general letter (replace \RL{آ ,أ , إ} with \RL{ا} ) \cite{c17} and removing the repeated letter used in reviews for intensification the opinions (\RL{رااااائع}). This technique was used in many works like \cite{c8,c11,c12,c13}.

\item Text filtering: removing words and symbols that do not have any effect on the output, like some stop words, punctuations and diacritics, etc. Almost all sentiment analysis researches have applied this technique like \cite{c5,c8,c9,c11,c12,c13,c16}.

\item Text stemming: replacing the word with its stem. This technique has a big impact in minimizing the required storage and in eliminating the redundant terms \cite{c17,c19} because there are many Arabic words have the same stem, and replacing the words with the same stem will reduce the redundant temrs, e.g the words \RL{تكره, أكره , يكره, يكرهون} have the same stem which is \RL{كره} , so all these words will be replaces with word \RL{كره}. The stemming technique was used in many works like \cite{c8,c9,c16}. Khoja stemmer \cite{c21} is an open source for stemming arabic text, and it has already used in many Arabic works like \cite{c64,c54,c101,c80,c28,c53}.

\item Emoticons converter: replacing the emoticon with their respective meaning in language. This technique was used in \cite{c8,c28,c80}.

\item Dialect converter: mapping the slang word to MSA. This technique was used in \cite{c8,c28}.

\item Negation and intensification: detecting negation and intensification phrases in dataset. This technique is very important for sentiment extraction in order to give the opposite polarity in case of negation and to give more weights to the current polarity in case of intensification. This technique was used in \cite{c8,c11,c12,c13}, e.g negation: \RL{هذا المطعم غير جيد} the word \RL{غير} will reflect the output polarity knowing that the word \RL{جيد} will get a positive polarity but the final output should be negative, intensification: \RL{هذا المطعم جيد جدا} the word \RL{جدا} will increase the weight of word \RL{جيد}

\item Part-Of-Speech (POS) tagging: Getting the type of the word such as noun, verb, adjective, etc, and give the word of these types more weight. This technique was used in \cite{c5,c13,c14,c15}. There are two available Arabic POS \cite{c22,c23}.

\item Feature extraction: For training machine learning classifiers, the annotated data must converted to feature vector, and a combination of specific features will give a specific class. The single feature is produced by converting a piece of text to a feature. In sentiment analysis studies, there are many features have been used, the most common features: Term Frequency, Term Unigrams, Parts Of Speech (POS), and Negation. And to make the job easier, many tools have been used in Sentiment Analysis researches like Rapidminer \cite{RapidMiner} as in \cite{c8,c9,c11}, MATLAB \cite{MATLAB} as in \cite{c24}, R \cite{R} as in \cite{c43}, and WEKA \cite{WEKA} as in \cite{c57,c103,c101,c90}.

\end{itemize}

\subsection{Sentiment classification}

\begin{figure*}
\centering
  \includegraphics[width=\textwidth]{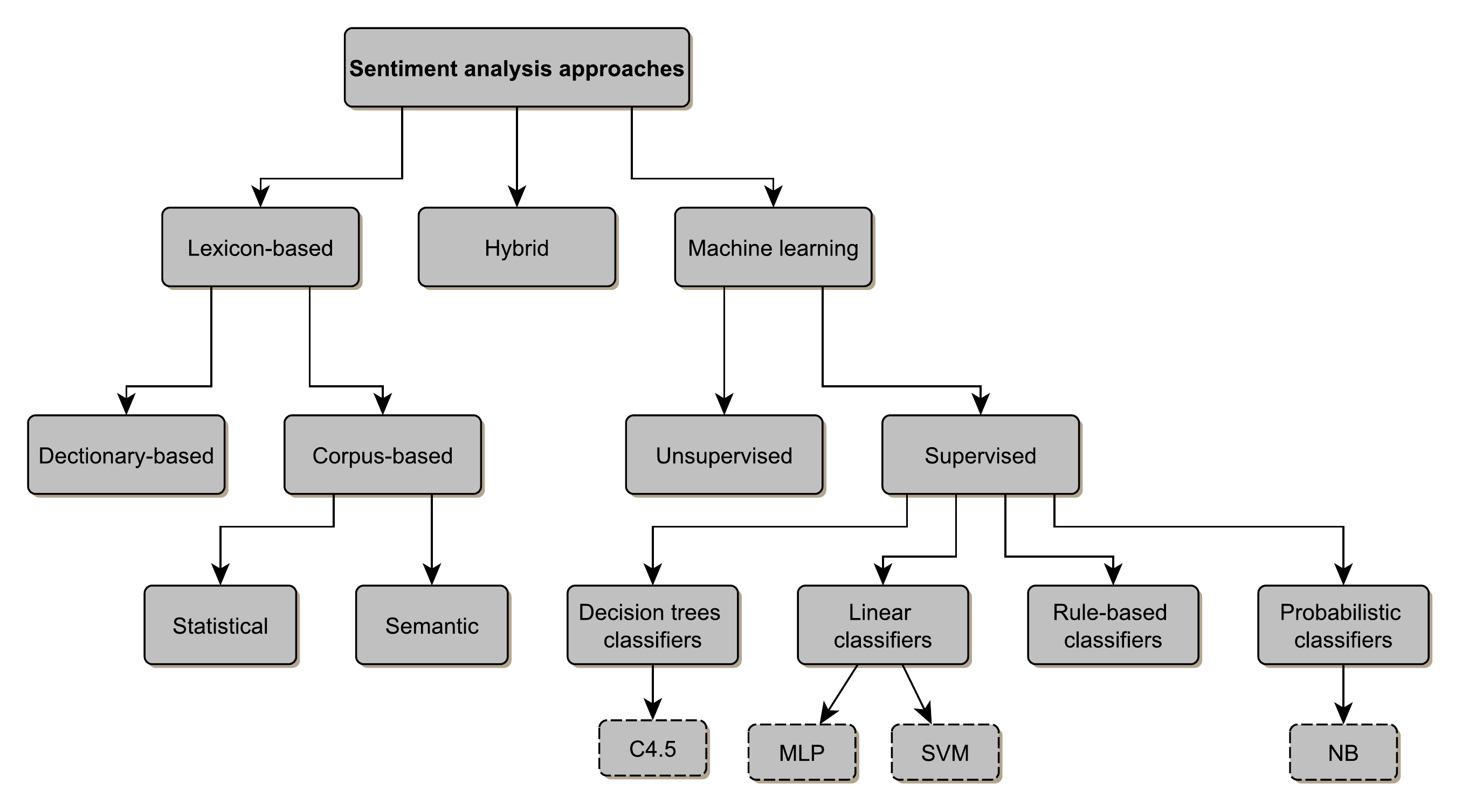}
  \caption{Classification of Sentiment Analysis Approaches.}
  \label{fig:SA_Approaches}
\end{figure*}

The third step in Sentiment analysis process depends on the used approach for classification. As mentioned in the introduction, sentiment classification approaches can be categorized into machine learning approach, lexicon-based approach, and hybrid approach which each of which can be categorized further into sub-categories as shown in Figure \ref{fig:SA_Approaches}. 

Table \ref{tab:Number_Approaches} shows the count of publications that adopted each approach for Arabic SA with a reference for each paper. As it can be seen, Most SA works for Arabic language used ML approach with 45 papers, followed by the lexicon approach with 25 papers, then the hybrid approach with only 12 papers. In the following subsections we describe each of the three approaches in details.

\begin{figure}
\centering
  \includegraphics[scale=0.4]{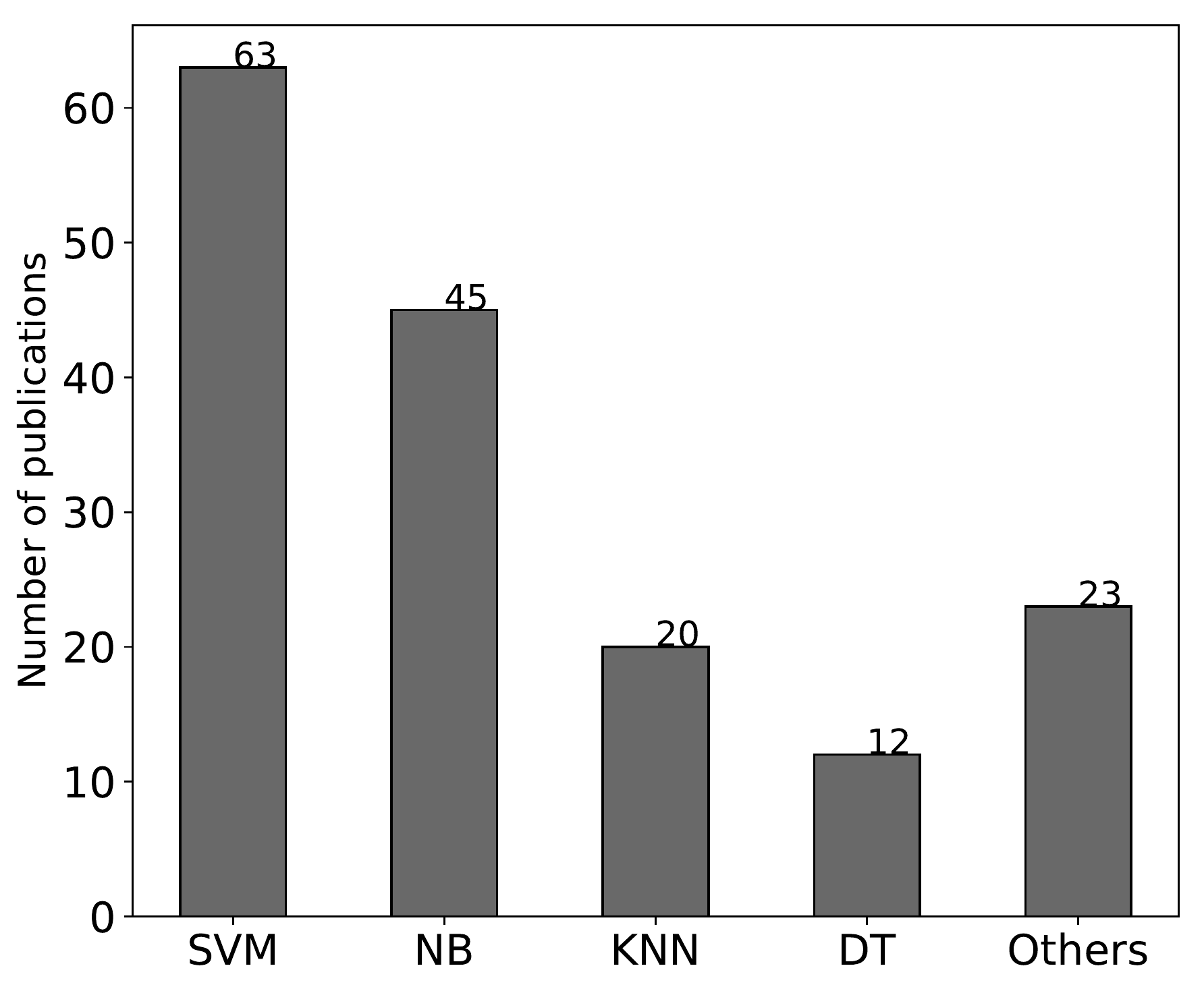}
  \caption{The count of publications for most applied machine learning classifiers in Arabic sentiment analysis}
  \label{fig:SA_Classifiers}
\end{figure}

\begin{table}
\caption{Sentiment Analysis approaches}
\label{tab:Number_Approaches}
\begin{center}
\begin{tabular}{|c|p{1cm}|c|}
\hline
Approach & Number of uses & Used in\\
\hline
ML & 45 & \multicolumn{1}{|p{4cm}|}{\centering \cite{c8}\cite{c9}\cite{c10}\cite{c27}\cite{c28}\cite{c29}
\cite{c32}\cite{c44}\cite{c50}\cite{c60}\cite{c63} \cite{c71}\cite{c72}\cite{c74}\cite{c76}\cite{c77}
\cite{c78}\cite{c79}\cite{c81}\cite{c82}\cite{c87} \cite{c88}\cite{c89}\cite{c90}\cite{c91}\cite{c92}
\cite{c94}\cite{c95}\cite{c97}\cite{c100}\cite{c101}
\cite{c102}\cite{c104}\cite{c106}\cite{c112}\cite{c113}
\cite{c114}\cite{c115}\cite{c123}\cite{c126}\cite{c127}
\cite{c128}\cite{c129}\cite{c11}}\\
\hline
Lexicon & 25 & \multicolumn{1}{|p{4cm}|}{\centering \cite{c12}\cite{c24}\cite{c36}\cite{c45}\cite{c46}\cite{c49}
\cite{c51}\cite{c53}\cite{c54}\cite{c58}\cite{c59}
\cite{c61}\cite{c67}\cite{c69}\cite{c70}\cite{c73}
\cite{c75}\cite{c83}\cite{c93}\cite{c96}\cite{c98}
\cite{c99}\cite{c11}\cite{c117}\cite{c125} }\\
\hline
Hybrid & 12 & \multicolumn{1}{|p{4cm}|}{\centering \cite{c13}\cite{c31}\cite{c33}\cite{c40}\cite{c41}\cite{c42}
\cite{c43}\cite{c47}\cite{c52}\cite{c57}\cite{c65}\cite{c107} }\\
\hline
\end{tabular}
\end{center}
\end{table}

\begin{table}
\caption{Dataset}
\label{tab:Used_Dataset}
\begin{center}
\begin{tabular}{|c|p{1cm}|l|}
\hline
Context & Number of Uses & Used in\\
\hline
Tweets & 45 & \multicolumn{1}{|p{3cm}|}{\centering 
\cite{c8}\cite{c11}\cite{c27}\cite{c28}\cite{c29}
\cite{c31}\cite{c43}\cite{c51}\cite{c53}\cite{c56}
\cite{c57}\cite{c58}\cite{c60}\cite{c61}\cite{c63}
\cite{c64}\cite{c68}\cite{c69}\cite{c70}\cite{c71}
\cite{c72}\cite{c78}\cite{c79}\cite{c81}\cite{c84}
\cite{c85}\cite{c12}\cite{c13}\cite{c42}\cite{c45}
\cite{c46}\cite{c54}\cite{c65}\cite{c66}\cite{c75}
\cite{c76}\cite{c77}\cite{c97}\cite{c99}\cite{c119}
\cite{c125}\cite{c126}\cite{c127}\cite{c128}
\cite{c129}}\\
\hline
OCA dataset\cite{c10} & 11 & \multicolumn{1}{|p{3cm}|}{\centering 
\cite{c10}\cite{c39}\cite{c52}\cite{c67}\cite{c94}
\cite{c96}\cite{c98}\cite{c99}\cite{c100}\cite{c101}\cite{c106} }\\
\hline
\multicolumn{1}{|p{2cm}|}{\centering Other (web forums, comments, reviews, etc.)} & 49 & \multicolumn{1}{|p{3cm}|}{\centering 
\cite{c9}\cite{c13}\cite{c24}\cite{c32}\cite{c33}
\cite{c40}\cite{c41}\cite{c42}\cite{c44}\cite{c45}
\cite{c46}\cite{c47}\cite{c48}\cite{c49}\cite{c50}
\cite{c55}\cite{c59}\cite{c62}\cite{c66}\cite{c74}
\cite{c75}\cite{c80}\cite{c67}\cite{c87}\cite{c88}
\cite{c90}\cite{c91}\cite{c62}\cite{c95}\cite{c97}
\cite{c102}\cite{c103}\cite{c104}\cite{c105}
\cite{c107}\cite{c110}\cite{c112}\cite{c113}
\cite{c114}\cite{c115}\cite{c116}\cite{c117}
\cite{c118}\cite{c120}\cite{c121}\cite{c122}
\cite{c123}\cite{c124}\cite{c129}}\\
\hline
\multicolumn{1}{|p{2cm}|}{\centering Dataset type is not mentioned} & 5 & \multicolumn{1}{|p{3cm}|}{\centering 
\cite{c30}\cite{c36}\cite{c73}\cite{c83}\cite{c93} }\\
\hline
\end{tabular}
\end{center}
\end{table}

\subsubsection{The machine learning approach}

The machine learning approach involves getting computers the ability to act without being programmed. Computer programs uses the exposed data to detect patterns and then adjust program actions and make intelligent decisions accordingly. In sentiment classification, this approach relies on the usage of famous machine learning techniques on the text. Most of researches on Arabic sentiment analysis used ML approaches because it was reported that they are more accurate than the lexicon approaches \cite{c203,c11,c108,c8}.

Machine learning can be classified into two main categories: unsupervised learning and supervised learning. In unsupervised learning, the learning algorithm does not require annotated documents and it will automatically discover some structure in the given documents. In contrast to unsupervised learning, in supervised learning, the algorithm will use the annotated documents for training. Supervised learning involves using already annotated documents and machine learning algorithm in order to train a supervised classifier. After preparing and annotating the dataset, it will be divided into training and testing datasets. Training dataset will be used for training the classifier and testing dataset will be used for testing and evaluated the classifier. 
In this training process, the classifier will learn from the annotated documents and this will allow it to make a prediction about the documents that might come in the future. The supervised learning is widely used for building sentiment analysis system and it can be categorized further into four types: Decision Tree Classifiers, Linear Classifiers, Rule-based Classifiers, and Probabilistic Classifiers.

Based on our survey, we found that the most used supervised classification algorithm in Arabic sentiment analysis is Support Vector Machine (SVM) which belongs to Linear Classifiers category. SVM is an algorithm that transforms training data into higher dimension by using a non-linear mapping function, and within the new dimension, it determine the best linear separator between different classes \cite{c202}. It was reported in many studies  studies that SVM outperforms other classifiers in Arabic SA classification like in \cite{c10,c11,c29,c31}.

Naive Bayes (NB) which belong to Probabilistic Classifier is the second applied ML classifier in Arabic SA. NB is a statistical classifier based on Bayes’ theorem, and it uses the probability theory in order to find the most likely class \cite{c202}. It is considered as one of the simplest classifiers. 
% The NB algorithm calculates the probability for each class and then the class with highest probability value will be selected \cite{c201}.

Figure \ref{fig:SA_Classifiers} shows the most commonly used classifiers for Arabic SA and the number of papers that applied each of them. As we can see, there are four main classifiers which are SVM, NB, KNN, and DT. SVM is most applied where 63 papers used SVM which forms around 39\% of the total number of papers that applied ML classifiers. NB comes next with 45 papers. K-NN and DT are less common with 20 and 12 papers, respectively. Other classifiers were also applied for Arabic SA in few studies like Conditional Random Field in \cite{c40}, Hybrid SVM \& KNN in \cite{c44}, Functional Tree in \cite{c47}, Genetic-Reduce in \cite{c85}, and Random Sub Space(RSS)\& SVM in \cite{c102}. All the other classifiers form around 14\% of all papers.

%%%%%%%%%%%%%%%%%%%%%%%

% Hossam: Why this paragraph is here ? How it is connected to the text before ?
%On the other hand, the authors in \cite{c8} used the ML approach, they developed a framework that analyze Arabic tweets and comments and then classify them to positive, negative, or neutral sentiment. They collected about 25000 tweets using PHP script that interact with Twitter’s Search API, the annotation was done by the crowd-sourcing technique. The built framework was able to analyze Arabic dialect by mapping dialect words to MSA. All experiments were built using RapidMiner, and three ML classifiers were applied: KNN, NB and SVM, the best accuracy was 76.78 \% and it achieved by NB classifier.
%Also, The authors in \cite{c29} have investigated the properties of the collected dataset and they have presented a set of benchmark experiments to their dataset in order to establish a baseline for future comparisons, they used many classifiers in their experiments like SVM, KNN, and NB, SVM was the best one.
%There are more articles can be added here. \cite{c27}

\subsubsection{The lexicon-based approach}

Lexicon-based approach depends on lexicon which contains a collection of sentiment words, each word has a polarity value, positive words have values greater than zero, negative words have values less than zero, and any word that does not exist in the lexicon is considered as a neutral word. For example a sentiment classification task can be performed based on this approach by looking for sentiment words in a given text or a document then adding weights or tags to these words and then counting the weights and tags to detect the overall sentiment. Sentiment words list with its polarity value exist in a sentiment lexicon, and to prepare the sentiment lexicon there are two approaches: dictionary-based approach and corpus-based approach \cite{c203}\cite{c204}\cite{c205}.

%\subsubsection{Dictionary Based approach}
Dictionary-based approach works as follows, starting from initial seed set of sentiment words with known positive and negative orientation. Then exploit available thesaurus and corpora like WordNet \cite{c206} to find synonyms and antonyms for each word in the list. The newly found word are added to the seed list and the next iteration starts. The process will have completed when no new words can be found, this approach is used in \cite{c207}\cite{c208}. The major disadvantage of this approach which that it does not find opinion words with domain orientation, for example, if we say: the phone speaker is quiet, this indicates a negative opinion, but if we say: the car is quiet, this indicates a positive opinion. 

%\subsubsection{Corpus Based approach}
Unlike the dictionary based approach, corpus-based approach can find domain and context specific opinion words. This approach rely on statistic or syntactic patterns with seed list of opinion words with known polarity to find new sentiment words with their polarity in a large corpus \cite{c203}\cite{c205}. For statistical pattern, the new sentiment word can be found by its occurrence frequency in a large annotated corpus, so if the word appears more frequently in positive documents than the negative documents then it will be added to words list as a positive word, and if it appears more frequently in negative document then it will be added as a negative word. In other words, the word will be added as a positive word if it occurs more frequently in a positive documents, and it will be added as a negative word if it occurs more frequently in a negative documents \cite{c209}. In syntactic pattern, the words with similar opinion appear together in the corpus, and this pattern suppose that if the words appear frequently together within the documents then they are likely have the same polarity \cite{c210}. So the word that does not have a polarity value and it appears frequently with another word with known polarity then it should have the same or opposite polarity value of the known word based on the connected word between them like word (AND), for example, the word (spacious) in this sentence (the car is comfortable \textbf{and} spacious) will have the same polarity value of the word (comfortable).

%%%%%%%%%%%%%%%%%%%%%

For constructing and expanding Arabic lexicon, there are two ways to do that in the literature \cite{c12,c13}: Manual and Automatic. In manual techniques, the lexicon is built by translating the content of SentiWordNet \cite{c6} or SentiStrength \cite{c25}, and for each translated word finds its synonyms and add the word with its synonyms to lexicon. In automatic technique, the lexicon is built by starting from manually collected and annotated lexicon (base lexicon) and then increase the lexicon size by adding synonym and antonyms. The manual method for lexicon creation is more accurate than the automatic one, but the automatic method required less time and work \cite{c12}, and according to the study \cite{c11}, the lexicon with bigger size has a better accuracy, so the lexicon should contain most possible sentiment words in order to increase the accuracy of the classifier that uses the Lexicon. 

%  paragraph to highlight the advantages of this approach
Lexicon-based approach is domain independent and the lexicon is built for all domains on the contrary of the ML approach which does not work well on data that differ from the training data \cite{c110}. Also, in this approach there is no need to have an annotated data for training process \cite{c9}.

An interesting example of a work that applied both approaches the corpus-based and the lexicon-based approach for Arabic SA was conducted by Abdulla \emph{et al.}  in \cite{c11}. In order to build the lexicon they collected 3479 negative and positive words, positive words are weighted with +1 value and negative words with -1 value. The lexicon was built in three phases, in each phase they increased number of words in the lexicon, and they proved that the bigger lexicon will get better results. The text was classified based on the total weights, if the total is greater than zero then it is positive, and if it is equal to zero then it is neutral and otherwise it is considered as a negative. The achieved accuracy using the lexicon was 58.6 \%. For the corpus-based approach, 2000 labeled tweets were collected by using a tweet crawler, 1000 positive and 1000 negative, and after applying the preprocessing steps they used four classifiers: SVM, NB, D-Tree, and KNN, The best result was 87.2 \% by using SVM classifier which is better than the lexicon based accuracy.

% Muath, this paragraph is very important. You can start it by talking about this disadvantages of this approach which were mentioned in the literature. Support that by citing the papers that mentioned these disadvantages or drawbacks. Then you can give some examples on how different works in the literature proposed some methods to overcome these problem. So lets start with something like this: Although the lexicon based approaches have their own advantages, however, they still suffer major drawbacks such as ... \cite{}, ... \cite{} and ...\cite{}. Therefore, many researchers proposed different methods and techniques to overcome these drawback. For example in .... the author proposed combing this and that ... \cite{}. Another example was presented in \cite{} where the authors proposed using .. this and that .. 
%Done

% This approach is much better than listing the works like In \cite{} , did this that from the first point to the last .. hope you get what i mean :)

% please follow the same approach of writing for the other sections when talking about advantages and disadvantages.

%After all, I think you are doing great. Please keep it up ;)
%Hossam

% 
Although the lexicon based approaches have their own advantages, however, they still suffer major drawbacks such as lack of availability \cite{c111,c24}, detecting the idioms \cite{c111,c46}, limited size \cite{c24,c108}, and requiring to a large amount of linguistic resources \cite{c110}. Therefore, many researchers proposed different methods and techniques to overcome these drawback. For example in \cite{c24} the authors proposed combining the available resources: English SentiWordNet (ESWN) \cite{c6}, Arabic WordNet (AWN) \cite{c34}, and the Standard Arabic Morphological Analyzer (SAMA) \cite{c35}, the combination has done by merging two lexicons in order to create a Large Scale Arabic Sentiment Lexicon (ArSenL), the first lexicon was created by matching AWN to ESWN and the second lexicon was developed by matching lemmas in the SAMA lexicon to ESWN, ArSenL has 28,780 lemmas and it is publicly available. Another example was presented in \cite{c46} where the authors presented a large scale Arabic idioms/proverbs sentiment lexicon of MSA and colloquial for sentiment analysis tasks, the idioms/proverbs lexicon was manually collected and annotated, it contained 3632 idioms and proverbs, the authors have proved that the idioms/proverbs lexicon can improve the sentiment classification process.

\subsection{The hybrid approach}
The hybrid approach combines both the machine learning and the lexicon-based approaches. This approach has been reported as a better approach than the machine learning and the lexicon-based approaches \cite{c43,c31,c110,c57}. 
The combination can be done by adding some extracted features from lexicon into machine learning classifier as well as other features such as sentence-level and linguistic features \cite{c13}. Another way to combine both approach is by removing each word that does not exist in the lexicon from all instances so only sentiment words will be remaining. The removed words do not affect the sentiment but keeping them will confuse the classifier and decrease the accuracy. Therefore, by removing these words the accuracy is expected to improved \cite{c43}. 

Compared with previous approaches, few works have used the hybrid approach for SA for Arabic language. For example, the work in \cite{c31} has presented a hybrid sentiment analysis approach for classifying Egyptian dialect tweets. In this work, the authors have combined the two sentiment analysis approaches, Machine learning and lexicon-based approaches. In Machine learning approach, three features were used: unigram, bigram and tira-gram. And for lexicon-based approach, they have collected the sentiment words from their corpus and gave each word a weight based on its frequency. The combination between the two approaches was done by adding the sentiment words as a features into Machine learning approach. And in the results, the accuracy of the hybrid approach was better than each single one.   

Another work in this research line was presented in \cite{c33}. This work presented a hybrid approach for Arabic opinion question answering by adding lexicon features to three classifiers, SVM, NB, and KNN. After preparing the data and applying different pre-processing techniques they have tested the three classifiers without lexicon features, and after that they have tested the same classifiers with lexicon features. In all experiments there was a remarkable enhancement by adding lexicon features to the three classifiers.

Another interesting work can be found in \cite{c57}. This work presented a hybrid sentiment analysis approach by removing all unsentimental words from training dataset so the training dataset contained sentimental words, instances and their label. The authors in this work have used two classifiers, SVM and KNN, and they have compared the proposed approach with the machine learning approach by using the same classifiers. The results have confirmed that the proposed approach has better accuracy than the machine learning approach.

%Few researches worked on this approach, for example in \cite{c13} the authors have presented a semi-supervised approach for sentiment analysis by adding some extracted features from lexicon to machine learning classifies (SVM) as well as other features such as sentence-level and linguistic features, two lexicons were used for improving the detection of the sentiment polarity, the first one is Arabic sentiment words lexicon which was built using manually collected and annotated words and then applying a new algorithm for expanding and detection the sentiment orientation of new words. The second lexicon is a phrase lexicon; this lexicon contained old wisdoms and popular idioms that were used by people’s comment to represent their opinions. The proposed model was applied on both MSA and Egyptian dialect using a corpus contained different types of data such as tweet, product reviews, hotel reservations, etc.

\subsection{Evaluation and assessment}
In the fourth step, the prepared module will be evaluated using one of the evaluation metrics like Accuracy, Precision, Recall, or F-score. Most studies have used tweets for experiments and testing, and other studies have used the dataset that used in \cite{c10}, it is available for the research community. Table \ref{tab:Used_Dataset} shows the statistical for the type of the used datasets in Arabic Sentiment Analysis studies.
\color{black}
\section{A proposed classification approach for Arabic Sentiment Analysis}

% Reviewer #1: The paper has given the survey in detail covering the three approaches used for sentiment analysis but it shall also incorporate a proposed approach also which the author believe he will use in future, along with the tool to be used.
%A model can be included in the paper to show the working of proposed approach. and how it would be different from existing discussed appracohes.

% %//// the added paragraph - Moath
% Based on our survey and because of the SVM is the most commonly used and the most accurate algorithm in Sentiment Analysis research, we proposed a Machine Learning approach which uses the SVM algorithm, and using the Grid Search in order to adjust the SVM parameters, gamma and C.
% Furthermore and to speed up the process, we would use the RapidMiner tool to build the proposed model because it contains on many already implemented modules like stemming, in addition to the ability to create or add extension modules. Moreover and unlike the most articles, we would focus on the restaurants reviews instead of tweets. 

Based on our survey on sentiment analysis for Arabic language, it can be noticed that there is an incremental trend toward the application of SVM as a machine learning classifier. The success of SVM algorithm for sentiment analysis specially for Arabic language is probably mainly due to its efficiency in terms of separability of text datasets with large number of features. The efficiency of SVM algorithm and its performance mainly depends on the kernel type and the selection of its parameters' values such as the Cost and Gamma.

In this section, we propose a general model based on SVM that can be used for feature selection and classification simultaneously. In addition, SVM's parameters are usually set manually or by using a grid search algorithm, which are both time consuming and inefficient processes. On the other side, Evolutionary Algorithms (EA) are considered as one of the fastest growing algorithms that can be used to solve complex optimization problems. EA are a type of stochastic search algorithms that are inspired by evolution theories in nature. EA showed very promising results in optimizing machine learning models in the literature \cite{faris2016mgp,faris2018improved,aljarah2016training}. We propose the application of these algorithms for optimizing SVM's parameters and locating the best features to build Arabic sentiment analysis classification model. 

\begin{figure*}
\centering
  \includegraphics[scale=0.45]{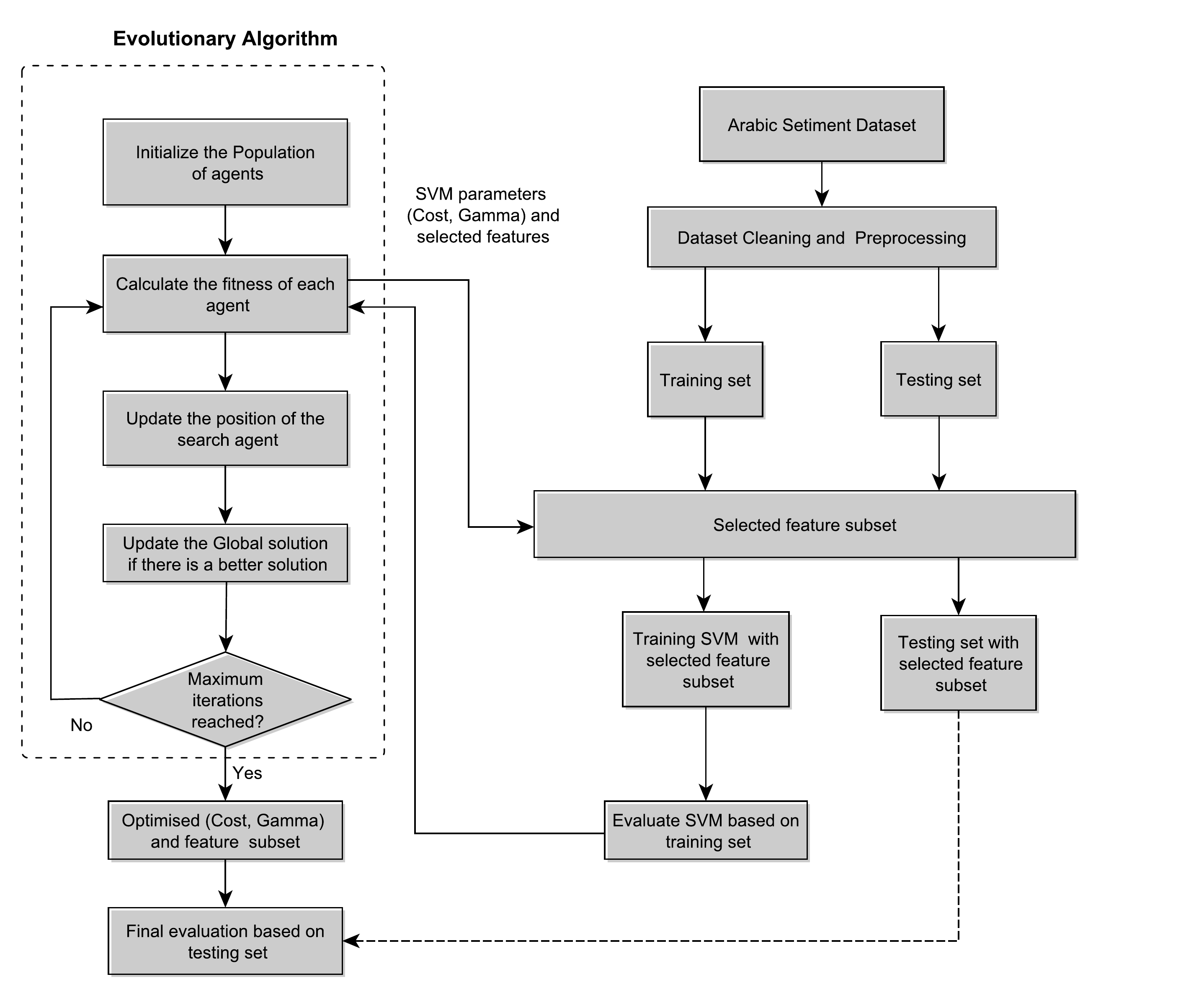}
  \caption{\color{black}A proposed classification approach for Arabic sentiment analysis based on SVM and EA.\color{black}}
  \label{fig:prop}
\end{figure*}

The following steps describe the proposed approach:

\begin{itemize}
\item Arabic sentiment dataset collection and preparation. 

\item Cleaning and preprocessing, which are applied on Arabic datasets. Such processes include text normalization and filtering, stemming, and feature extraction.

\item Partitioning the dataset into two parts: training part and testing part. 

\item After that, the optimization process is started using any well-regarded evolutionary algorithm to search for the best SVM parameters and the most relevant feature subset for the SVM model based on the training part of the data based on a system architecture as described in \cite{aljarah2018simultaneous,faris2017multi}.

\item After an iterative process, the final SVM classification model which contains the best parameters and best features subset is evaluated based on the testing part using the sentiment analysis evaluation measures. The proposed classification approach for Arabic sentiment analysis can be described as shown in Figure \ref{fig:prop}
\end{itemize}

As we have seen in this survey, most of the previous Arabic sentiment analysis approaches applied standard machine learning algorithms without any modifications. In our proposed approach, we selected the best performing SVM classifier and suggested a mechanism to enhance its efficiency using very powerful search algorithms that proved their success in solving optimization problems. Moreover, the selected features after performing the optimization process can be further analyzed to study their impact in the Arabic sentiment classification problems. For implementation, we plan to utilize EvoloPy as a framework for development. EvoloPy is a simple open source optimization framework written in Python \cite{faris2016evolopy}. The framework contains a number of recent metaheuristic algorithms.

% In this section we need to propose a model as suggested by the reviewer

\color{black}

\section{Conclusions and possible research avenues}

Arabic sentiment analysis is considered one of the growing fields that concern with uncovering the feelings of the individuals toward specific event, brand, or something else. It is clear that the field of Arabic sentiment Analysis is in the early stage and the research on this field has rapidly increased over the last few years.

In this survey, we have investigated the Arabic sentiment analysis, and made comprehensive literature review about this important topic. All the previous studies are categorized into three main approaches: machine learning approach, lexicon-based approach, and hybrid approach. We found that most of researchers have focused on machine learning approach by applying different classification models such as SVM, and NB, which are the most used algorithms in the literature. The main disadvantage of the machine learning approaches is the domain dependent, which does not work well on data that differ from the training data. On the other hand, lexicon based approaches resolve this issue, where there no need for training data and it is considered a domain independent. Lexicon approach also has some drawbacks such as lack of availability, detecting the idioms, limited size, and requiring to a large amount of linguistic resources. Moreover, there is no stemming algorithms for dialect words of Arabic and Arabic dialects has more stopwords than the standard.

To resolve the shortcomings of the previous approaches, some researchers proposed hybrid approaches, which combine machine learning and lexicon approaches by adding some extracted features from lexicon into machine learning classifiers or by removing all words from the dataset, where these words do not exist in the lexicon. This approach has been reported as a better approach for Arabic Sentiment Analysis. In spite of this advantage, the number of publications that investigated the hybrid approach for Arabic language is much less than the other approaches. 

Regarding the lack of available datasets, most of works used tweets to do their experiments, and other works have used OCA dataset \cite{c10}.  Therefore, the research on SA for Arabic language lacks the publications related to other contexts and social networks. 

WordNet, SentiWordNet, and  Multi-Perspective Question Answering (MPQA) \cite{c211} are the most common linguistic sources used in English Sentiment Analysis, and building linguistic resources for Arabic language is still needed.  

% What are some promising areas to investigate?
% what is still missing in the arabic sentiment analysis work?
% Are there any interesting technologies or approaches applied for other languages and but not yet for the arabic case?

% we can say also that most of researchers have focused on this approach but less for that .. please elaborate

\bibliographystyle{ieeetr}
%\bibliography{references}

\end{document}